\documentclass[10pt,journal,compsoc]{IEEEtran}
\ifCLASSOPTIONcompsoc
  \usepackage[nocompress]{cite}
\else
  \usepackage{cite}
\fi
\ifCLASSINFOpdf
\else
\fi

\usepackage{cite}
\usepackage{amsmath}
\usepackage{multirow}
\usepackage{amsmath,amssymb,amsfonts}
\usepackage{algorithmic}
\usepackage{graphicx}
\usepackage{textcomp}
\usepackage{xcolor}
\usepackage{enumitem}
\usepackage{float}
\usepackage{graphicx,subcaption}
\usepackage{color}
\usepackage{float}
\usepackage{ragged2e}

\begin{document}
%
\title{Use of Transfer Learning and Wavelet Transform for Breast Cancer Detection}
%
%
%
%

\author{Ahmed~Rasheed,
        Muhammad~Shahzad~Younis, Junaid~Qadir~and~Muhammad~Bilal
\IEEEcompsocitemizethanks{\IEEEcompsocthanksitem Ahmed~Rasheed and Muhammad~Shahzad~Younis are with School of Electrical Engineering and Computer Science, National University of Sciences and Technology, Islamabad, Pakistan.
\protect\\
E-mail: arasheed.msee17seecs@seecs.edu.pk\hfil\break E-mail: muhammad.shahzad@seecs.edu.pk
\IEEEcompsocthanksitem Junaid~Qadir is with Information Technology University (ITU)-Punjab, Lahore, Pakistan.\protect\\
E-mail: junaid.qadir@itu.edu.pk 
\IEEEcompsocthanksitem Muhammad~Bilal is with University of the West of England (UWE), Bristol, England.\protect\\
E-mail: muhammad.Bilal@uwe.ac.uk
}
\thanks{}}

\IEEEtitleabstractindextext{%
\begin{abstract}
\justifying
Breast cancer is one of the most common cause of deaths among women. Mammography is a widely used imaging modality that can be used for cancer detection in its early stages. Deep learning is widely used for the detection of cancerous masses in the images obtained via mammography. The need to improve accuracy remains constant due to the sensitive nature of the datasets so we introduce segmentation and wavelet transform to enhance the important features in the image scans. Our proposed system aids the radiologist in the screening phase of cancer detection by using a combination of segmentation and wavelet transforms as pre-processing augmentation that leads to transfer learning in neural networks. The proposed system with these pre-processing techniques significantly increases the accuracy of detection on Mini-MIAS.

\end{abstract}

}

\maketitle

\IEEEdisplaynontitleabstractindextext

%
\IEEEpeerreviewmaketitle

\IEEEraisesectionheading{\section{Introduction}\label{sec:introduction}}

%
%
%
%
\IEEEPARstart{B}{ased} on the research of World Health Organization (WHO), 22.9\% of the total cancer cases diagnosed are of breast cancer and accounts for 13.7\% cancer-related deaths all over the world \cite{b1}, which places breast cancer at second place in the leading causes of cancer-deaths among women behind heart diseases \cite{b16}. 
Early detection of tumors or irregular masses can save the lives of many. Breast cancer has over 90\% chance of being cured at its early stage but it does not get much attention until it gets severe as it is not painful in its early stage.

Computer-aided systems are now of great interest to reduce the number of cancer casualties by detecting it in the early stages and helping out the radiologists in making important decisions \cite{b2}. There are two types of mammography examinations, screening and diagnostic. Diagnostic examination is a follow-up method for patients who are already classified with abnormal tissues by a proper channel in early screening process \cite{b3}. Screening of mammography generally has four views---the Craniocaudal (CC) view and Mediolateral (MLO) view for each breast. For further diagnostic more views are also referred for in-depth analysis by radiologists. In our system we worked on the classification of screening problem which will help the radiologists in correct diagnostics and determining the next steps.

We propose a system that utilizes the knowledge and techniques of image processing and deep learning resulting in the classification of masses found in the x-ray scans of breasts known as mammograms. The abnormalities are classified in the different classes based on the structure, shape, and composition of tissues present. Furthermore, the system can also tell the severity of the abnormality in two more classes (i.e., is the abnormality benign and malignant). In particular, Deep Learning and Image Processing techniques are used in our proposed system to classify the abnormalities in the mammographic scans obtained from ``Mammographic Image Analysis Society'' (MIAS) database.

In this work, we use a combination of Wavelet transforms, Segmentation and Convolutional Neural Networks (CNN) for breast cancer abnormality detection, which provides promising results with increased efficiency of detection. CNN is best for learning different features from each class under observation. It uses filters of different sizes which helps the network in learning and recognizing the patterns in the images \cite{b24}. Wavelet transform and segmentation are widely used image processing techniques that enhance the features present in the image scan. These features are then learned by the CNN to classify the images. Wavelet transform takes the image in frequency domain and then decompose it into sub-band images from which different features can be extracted easily \cite{b21} \cite{b22}. Segmentation on the other hand enhances the boundaries and corners in the image, it splits the the image into several portions. Each portion signifies a useful data in the form of different gray-levels. Segmentation of medical images highlight the important parts under study like the tumors in the problem under consideration \cite{b23}. The combination of wavelet transform and segmented image used as the inputs of the CNN helps the network in learning the significant features and improves the overall efficiency of the classification done by the network \cite{b25}. This study can further be used to develop Computer-aided Design (CAD) which would be useful for the radiologists to detect cancer in its early stages and take the necessary actions needed.


Training deep CNNs from scratch is not easy as a very large dataset along with high-end computing resources are needed. One way to resolve this constraint is to use \textit{Transfer learning} in which already-trained deep ML models are used in a related but different setting and tuned according to one's needs through the removal or addition of a small number of new layers at the end of the network and then training of only the newly added layers \cite{b15}. 
This method is especially effective when we have smaller datasets and wish to augment the learning on these datasets with more general readily-available information captured by the existing models \cite{b20}. Pre-trained Deep Neural Networks (DNNs) are tested on the database using transfer learning techniques and pre-processed images are passed to them. 

The proposed system presents a state of the art architecture which combines the knowledge of image processing and neural networks to form a single network in order to detect the breast cancer from the mammograms. Moreover, the system proposed does two step image classification. The first one, a seven class classification detects the abnormality in the image mammogram and the latter one, a three-class classification, tells the severity of the case using the mammogram and information obtained from the first network combined. The  arrangement forms a hybrid system which in turns improves the results of the classification by a significant factor and provides state of the art results for a three-class classification of mammograms.Breast cancer classification has been majorly accomplished via binary classification. \cite{b42}\cite{b43}\cite{b44} The proposed methodology goes a step further and uses three class-classification for breast cancer detection where, in addition to detecting the tumor, it also identifies whether the detected tumor is benign or malignant.

The rest of the paper is organized in the following way. Background and related works are presented in Section II. The details of the dataset are presented in Section III. The details of our methodology are provided in Section IV. Our results are provided in Section V.  Finally, the paper is concluded in Section VI.

 

\section{Background and Related Work}

Deep learning models are trained to achieve human-like intelligence and decision making. Lots of work already has been done in medical field to achieve models with state-of-the-art accuracy on different medical datasets and different problems \cite{b4}\cite{b5}\cite{b11}. Use of deep learning models can be traced back to late 90s. CNN architecture is widely used for the classification of image-based data. Already trained networks can be found on databases to classify the images into unique classes. Significant amount of work has already been done in classifying the mammograms based on their abnormalities. For classification various techniques are used for image pre-processing and augmentation to aid the neural network and improve its performance.

The background and related work done in medical imaging and most importantly in breast cancer detection has utilized a lot of  techniques however separately. Most breast cancer detection techniques have utilized wavelets for feature enhancement\cite{b27}\cite{b29}, segmentation for diagnosis\cite{b4}\cite{b46} and transfer learning CNN networks for classification.\cite{b47}\cite{b48}\cite{b49} in their individuality. The proposed system differs as it combines all three of these techniques and consequently gives a much improved results than achieved before with the individual techniques. The resultant AUC [Area Under the Receiver Operating Characteristic (ROC) curve] gives a much improved classification. 

\subsection{AUC-ROC}
Area Under the Curve - Receiver Operating Characteristic (AUC-ROC) Curve is a metric for establishing the extent to which a model can distinguish classes. AUC-ROC curves have been used as an apt measurement standard for medical imagery \cite{b41}.Higher the AUC-ROC, better the performance of the model. Our system generates an AUC-ROC curve, on the mini-MIAS dataset for breast cancer detection, that shows more promising results than previous works done.

\subsection{Segmentation}
The first technique used by our proposed system is image segmentation based on the intensities of the mammogram images. The segmentation helps the system to find the Region of Interests (ROI). \newline
For this purpose, noise is removed first by adaptive mean filtering as this filter removes the noise effectively and is better among noise removal filters. Adaptive filter works by comparing each pixel of the image to its neighboring pixel. If the pixel does not match with the majority of its neighboring pixels then it is considered as noise and is then replaced by the mean value.\newline
After the noise removal K-mean clustering algorithm is applied to the image. This algorithm selects the k number of centroids for the regions and then allocates the pixels to the regions based on the clustering of neighboring pixels. It is widely used in the segmentation of medical images as it is important to extract only the desired regions from an image which may not be easily distinguishable \cite{b34} \cite{b36}. For the classification purposes segmentation is used as an aid to improve the performance. BichenZheng \cite{b35} used a hybrid network of K-means segmentation and Support Vector Machine (SVM) to show the improved accuracies in the detection of breast cancer. 

\subsection{Wavelet Tranform}
Wavelet transform is a well-known image processing technique to extract the features from an image. This technique is used to augment the available data by providing the wavelets of original and segmented mammographic scans. The main advantage of using wavelets is that they do a simultaneous localization in frequency and time domain and is faster than other methods like FFTs \cite{b13} \cite{b17} \cite{b18}. 

Wavelet transform has been used in several researches to aid radiologists in understanding the image scans \cite{b26}\cite{b27}\cite{b40}. It enhances the features in the image scan which are also beneficial for neural networks to learn and classify the images more effectively. In medical imaging wavelets are mostly used as pre-processing technique for image feature extraction. Liu et al. \cite{b28}, Ferreira and Borges \cite{b29} and Rashed et al. \cite{b30} showed that using wavelet transform in classification improves its results in medical imaging.

Wavelet transforms helps to enhance the important features like the edges of tumor present in the breast scan.
The general equation of the wavelet transform is provided next.

\begin{equation}
    W_k^j=\int f(x)\Psi(\dfrac{x}{2^j}-k) dx
\end{equation}

\begin{figure}[h]
\centerline{\includegraphics[width=1\columnwidth]{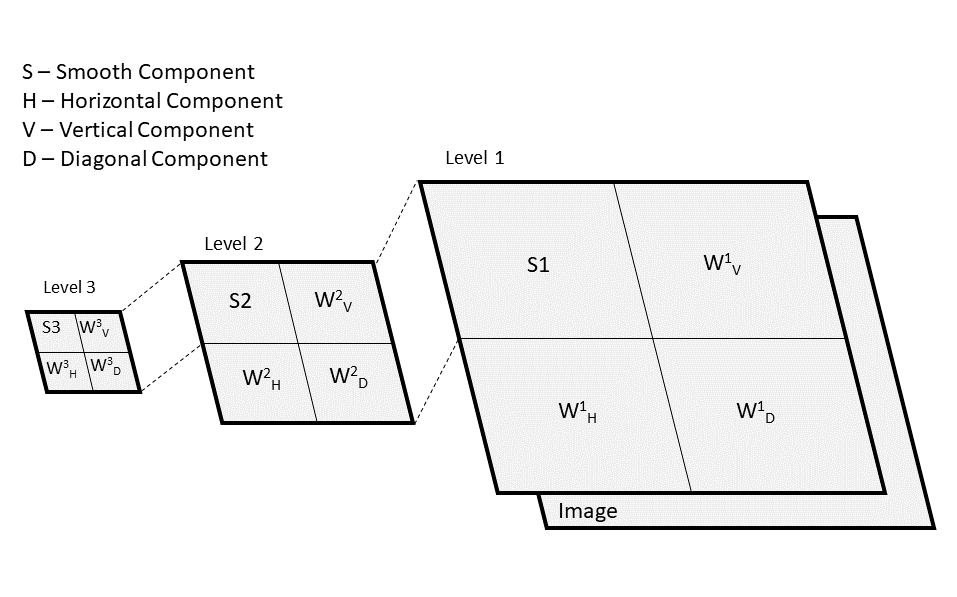}}
\caption{2D Wavelet transform layer representation up to 3 levels.}
\label{layers}
\end{figure}

Here $\Psi$ is transforming function, \textit{f(x)} represents the original signal while \textit{k} and \textit{j} are translation and scale parameters respectively \cite{b14}. 
Figure \ref{layers} shows the visual representation of 2D wavelet transforms. Here \textit{$W_D^j$}, \textit{$W_H^j$} and \textit{$W_V^j$} presents detailed components in diagonal, horizontal and vertical direction at the desired level \textit{j}. 

\subsection{Augmentation}
Dataset augmentation refers to the task of generating new images from the existing dataset, which improves the data scarcity and prevents the network from over-fitting  \cite{b8}. Image augmentation is used as the data enhancement technique when you do not have a big database of images. It is considered an important aspect when using CNN for classification as it improves the generality of the network \cite{b31} \cite{b32}. Holger R. Roth \cite{b33} proved that data augmentation significantly improves the performance of the network in classification problems.

\subsection{Deep Neural Network (DNN)}
DNN is used at the end for the classification among selected classes. DNN is now widely used in medical classification. A lot of work has also been done in classifying the breast cancer using DNN \cite{b37} \cite{b38} \cite{b39}.  \newline
These layers are then trained on the training dataset while the weights of all other layers in the pre-trained models are kept frozen. This technique is known as ``Transfer Learning''. The purpose of using transfer learning is to effectively utilize the complex structure of premade CNNs and to efficiently train the network on a smaller dataset. 



\begin{enumerate}[wide=0pt, label=\bfseries\arabic*.]

\item \textit{Pre-trained CNN:}  An already trained network is more useful instead of building a whole new CNN architecture from the scratch and then training it on our smaller dataset. This would not give good results on the output of the network. For this purpose, the number of already trained models are used by replacing the final three layers of these models with the new fully connected and classification layers according to our desired number of classes.

For the purpose of transfer learning, our proposed system uses the previously trained state-of-the-art CNNs. 

\begin{itemize}
\item \textit{AlexNet:} Krizhevsky in 2012 \cite{b6} achieved 15.3\% top-5 test error rate in the ImageNet Large Scale Visual Recognition Challenge (ILSVRC) with their unique CNN architecture called AlexNet that consisted of convolutional layers, dropout layers, pooling layers, and at last fully connected layers \cite{b12}. The success of AlexNet ushered in the era of deep learning and CNNs.

\item \textit{VGGNet:} Simonyan and Zisserman in 2014 introduced a 19-layer deeper CNN achieving top results in ImageNet ILSVRC \cite{b7}. Their proposed network had small convolutional filters which in the end showed a notable increase in accuracy. This work led CNNs architectures to have deep network of layers for better feature representations. 

\item \textit{GoogleNet:} Szegedy et al. in 2014 entered ImageNet ILSVRC with a deeper network. This network was trained on high end resources and achieved a 6.7\% error rate in top-5 test \cite{b8}.
This network used a parallel inception network instead of sequentially stacking the layers one after the other in the architecture. 

\item \textit{ResNet:} He et al. in 2015 introduced a network architecture consisting of 152 layers \cite{b9}.
This set a new record in ILSVRC by achieving a 3.57\% top-5 error rate in the classification of the ImageNet database. They introduced a huge number of smaller layers which made the model a very deep network but doing so reduced the complexity of it.
\end{itemize}

All the aforementioned deep network architectures have been trained on the ImageNet database and have been made designed to classify the input image into one of 1000 classes. To use them for the proposed problem, the concept of transfer learning is used. In transfer learning, the last three or more network layers are removed and new layers are appended with the desired number of neurons in the last fully-connected layer based on the number of classes for the desired task. Learning rate is set high for the newly added fully-connected layers. The network above remains unchanged and the new layers are trained on the newer dataset. A validation check is performed after every few iterations to check the performance of the network training.

\item \textit{Augmentation:}  The above-mentioned techniques, segmentation, and wavelet transforms, are also used as data augmentation and fed to the network for the training purpose. Apart from this, the basic augmentation is also done which includes the rotation, translation, scaling, and sheer of the images with random values.

\item \textit{Transfer Learning:} Transfer learning the transfer of knowledge obtained from one domain to another domain of different application. Transfer learning is a part of deep learninng and comes in handy when the data on which the system is to be trained is a small database \cite{b15} \cite{b19}. In the proposed system the it is used by transferring the knowledge of a pre-trained network to our setting which is abnormality detection in mammogram scans. 

Fig. \ref{learnpro} shows the visual representation of transfer learning process.

\begin{figure}[htbp]
\centerline{\includegraphics[width=.8\columnwidth]{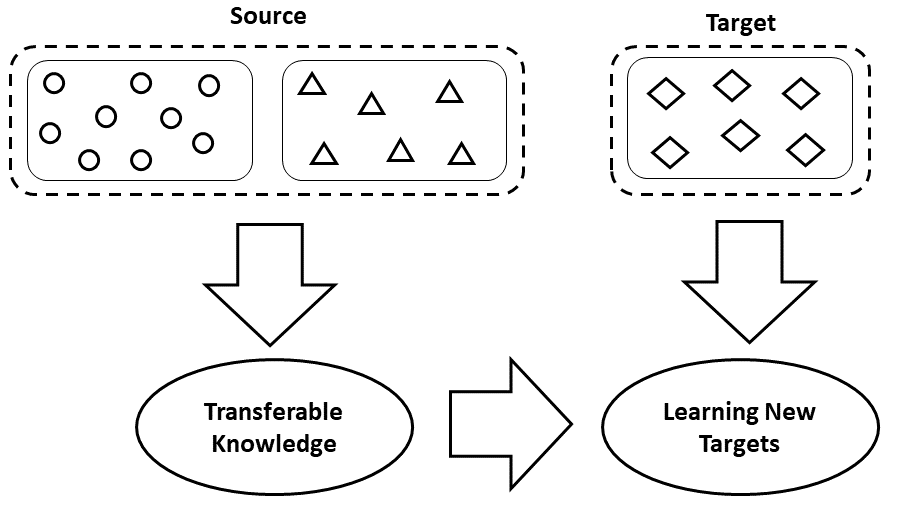}}
\caption{Transfer Learning Process}
\label{learnpro}
\end{figure}

\end{enumerate}

Our system proposes a novel structure using the combination of above mentioned techniques. A lot of work has been done on the given problem of classification using these techniques individually. Our proposed system made with the combinations shows a significant improve in the efficiency. The segmentation and wavelet transforms act as the pre-processed augmentation of image data which helps out the DNN in learning features in a more effective way. Furthermore, the proposed system uses two DNNs in a cascaded way, reported in the section below, which also helps the system learn the classes in a better way.


\section{Dataset}
Neural networks require a large number of images as input for the training purposes. Because of limited resources of large dataset availability, the dataset used for this research is mini MIAS dataset \cite{b10}. It comprises of total 322 images which are further divided into a number of classes. Original size of each image in the dataset is 1024$\times$1024. Sample images are shown in Figure 1. 

\begin{itemize}
\item \textit{\textbf{CALC}---Calcification}: deposition of calcium carbonates around the tissues (34 images).

\item \textit{\textbf{CIRC}---Well-defined/Circumscribed masses}: irregular shaped masses present in breast (24 images).

\item \textit{\textbf{SPIC}---Spiculated masses}: poorly defined masses in the form of abnormality (24 images).

\item \textit{\textbf{MISC}---Other, ill-defined masses}: miscellaneous abnormalities (18 images).

\item \textit{\textbf{ARCH}---Architectural Distortion}: abnormal tissue arrangement causes architectural distortions in the breast tissues (12 images).

\item \textit{\textbf{ASYM}---Asymmetry}: increased mass density of the breast tissue (21 images).

 \item \textit{\textbf{NORM}} represents a normal image devoid of any abnormalities. (189)

\end{itemize}

\begin{table}[H]
\centering
\captionsetup{justification=centering, labelsep=newline}
\caption{322 cases of Mini-MIAS dataset classified into 7 abnormalities.}
\begin{tabular}{|l|l|}
\hline
Classes & Number of Images \\ \hline
CALC    & 34               \\
CIRC    & 24               \\
SPIC    & 24               \\
MISC    & 18               \\
ARCH    & 12               \\
ASYM    & 21               \\
NORM    & 189              \\ \hline
Total   & 322              \\ \hline
\end{tabular}
\label{class7}
\end{table}

These classes are further divided into the severity of the abnormality present among the tissues based on the mammogram scans. Severity is divided into two classes.

\begin{itemize}
    \item Benign
    \item Malignant
\end{itemize}

\begin{table}[H]
\centering
\captionsetup{justification=centering, labelsep=newline}
\caption{322 cases of Mini-MIAS dataset classified into 3 classes}
\begin{tabular}{|l|l|}
\hline
Class     & Number of Images \\ \hline
Benign    & 67               \\
Malignant & 54               \\
Normal    & 201              \\ \hline
Total     & 322              \\ \hline
\end{tabular}
\label{class3}
\end{table}

Table \ref{class7} and \ref{class3} shows the exact number of image distribution among 7 classes of abnormality and 3 classes of severity respectively.

For training and testing purposes the dataset is further divided into two subsets: training data and validation data. 75\% images from each class are allocated for training of the network while the remaining 25\% are allocated for the validation of the trained network. This makes 90 images for validation and 232 for training. \par From 322 images, 133 images are of cases with abnormalities present and 189 images are of normal cases. Classes are divided in the following way as per the abnormality present. This splitting of dataset is totally random to ensure unbiased network training. The classes are also balanced at through random selection in order to get unbiased classification for both the networks while training. 


\begin{figure}[H]
  \begin{subfigure}[t]{0.16\textwidth}
    \includegraphics[width=\textwidth]{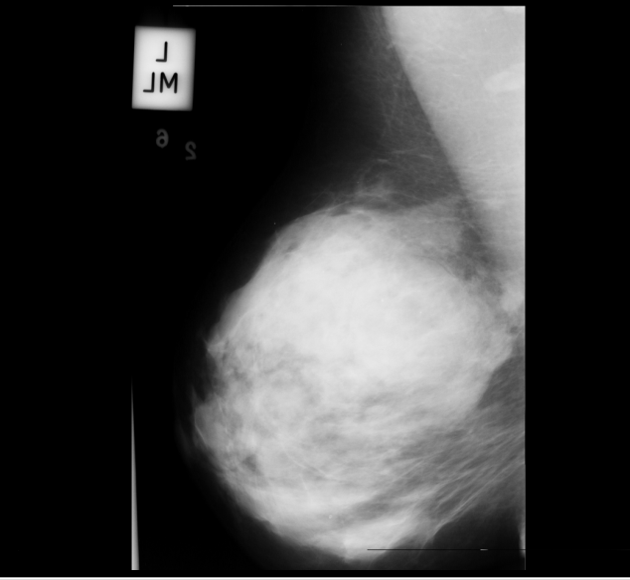}
    \caption{}
  \end{subfigure}\hfill
  \begin{subfigure}[t]{0.16\textwidth}
    \includegraphics[width=\textwidth]{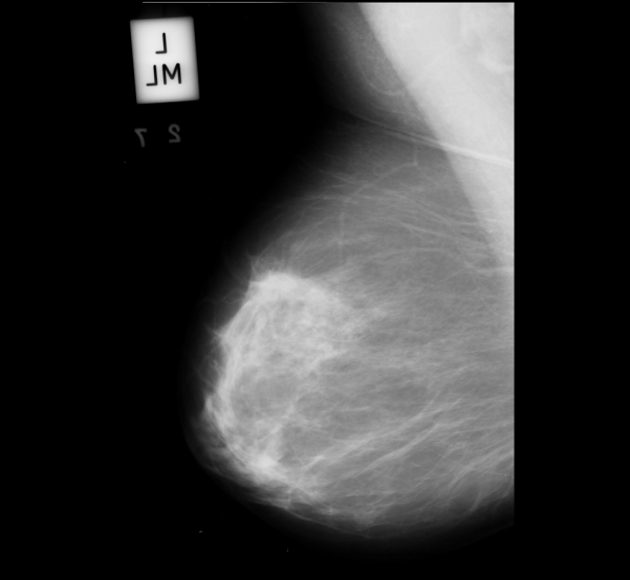}
    \caption{}
  \end{subfigure}\hfill
  \begin{subfigure}[t]{0.16\textwidth}
    \includegraphics[width=\textwidth]{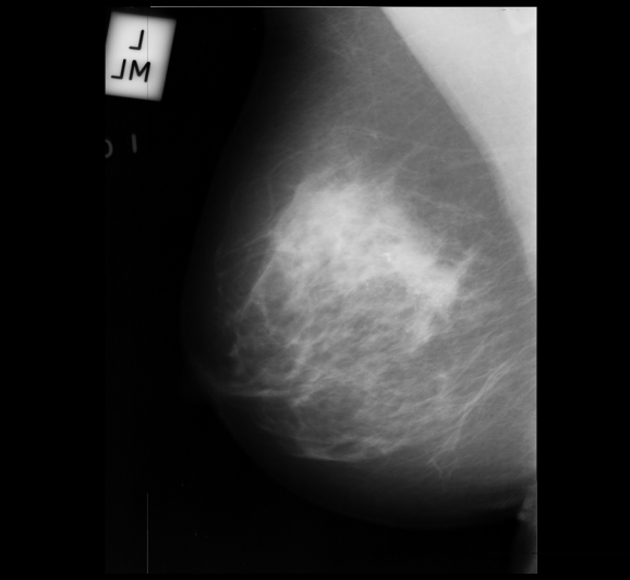}
    \caption{}
  \end{subfigure}\hfill
  \begin{subfigure}[t]{0.16\textwidth}
    \includegraphics[width=\textwidth]{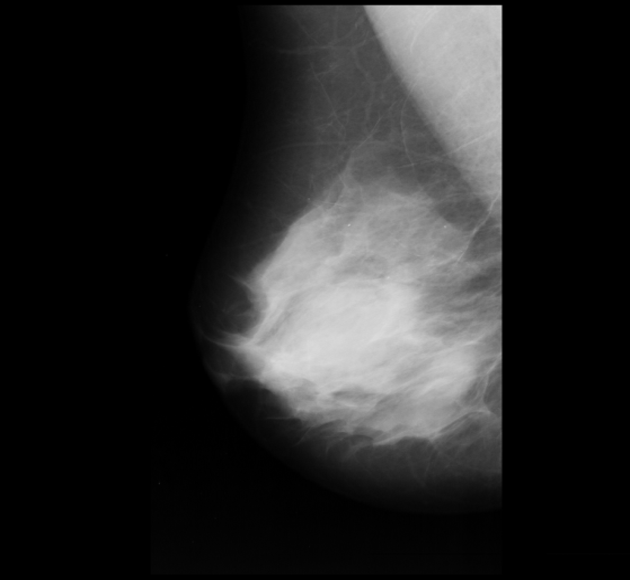}
    \caption{}
  \end{subfigure}\hfill
  \begin{subfigure}[t]{0.16\textwidth}
    \includegraphics[width=\textwidth]{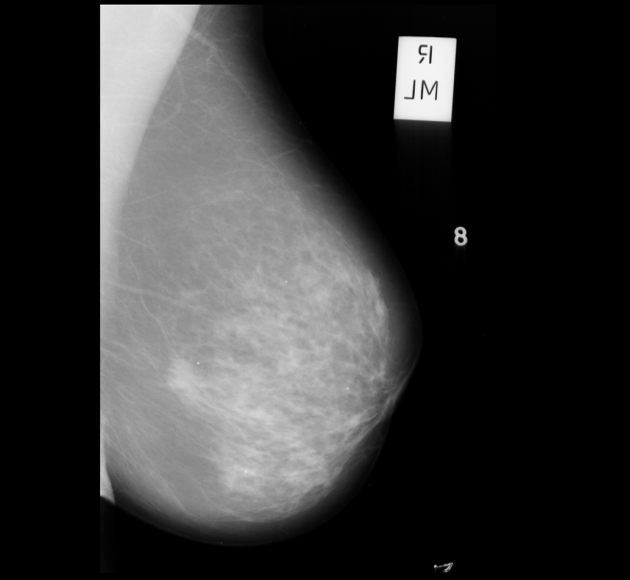}
    \caption{}
  \end{subfigure}\hfill
  \begin{subfigure}[t]{0.16\textwidth}
    \includegraphics[width=\textwidth]{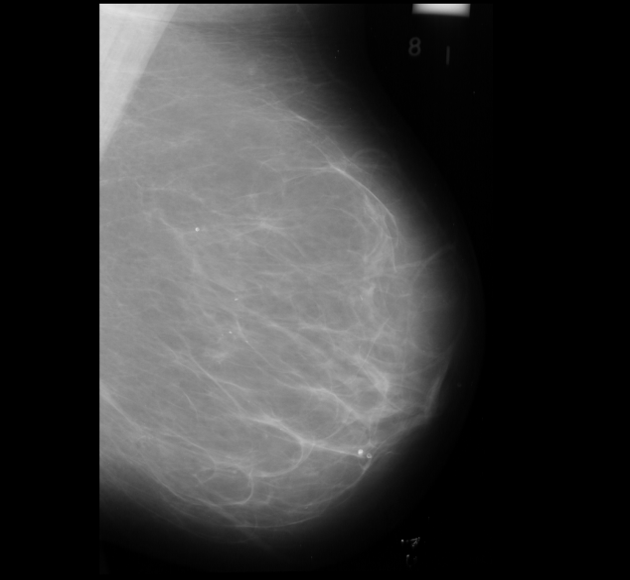}
    \caption{}
  \end{subfigure}
\caption{Sample images from the mini MIAS database. (a) ARCH (\textit{architectural   distortion}),  (b)  ASYM (\textit{asymmetry}),  (c)  CALC (\textit{calcification}),  (d)  CIRC (\textit{circumscribed masses}),  (e)  MISC (\textit{miscellaneous/ill-defined masses}), and  (f)
NORM (\textit{normal image devoid of masses}).}
\end{figure}


\section{Methodology}\label{AA}

\begin{figure}[H]
\centerline{\includegraphics[width=1\columnwidth]{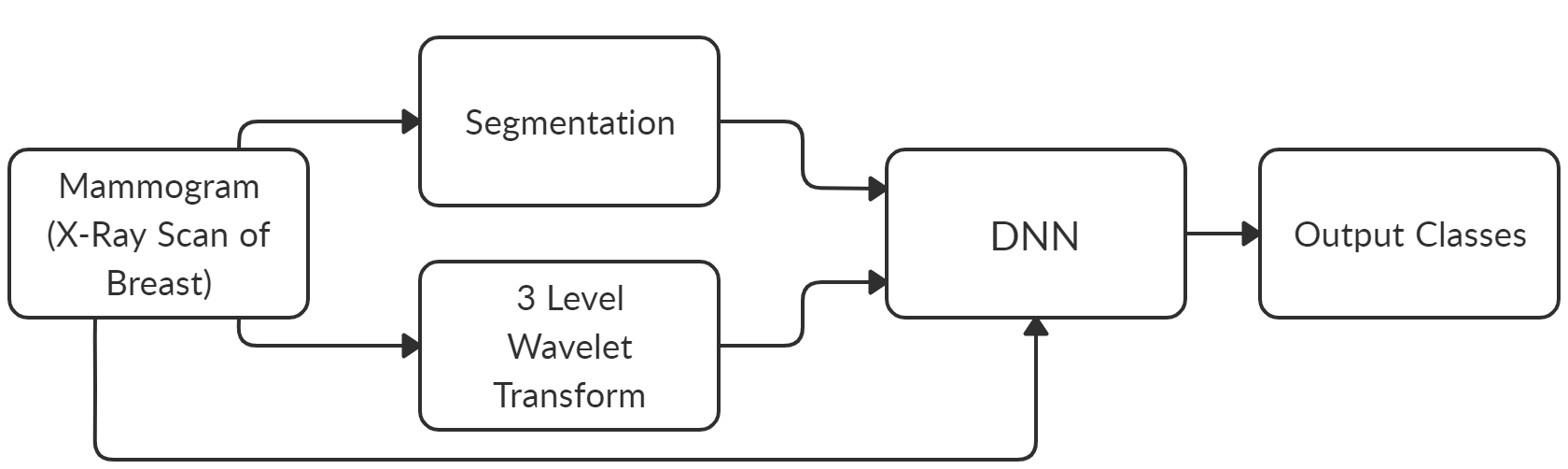}}
\caption{Proposed system workflow diagram.}
\label{workflow}
\end{figure}

The flow diagram in Figure \ref{workflow} illustrates our overall methodology.

\subsection{Segmentation}
The first step done is image segmentation based on the intensities of the mammogram images. K-means algorithm is used with adaptive noise removal to obtain the best results of segmentation. Image scan of mamogram is divided into portions and each portion is represented with a different gray level intensity. These different intensities extracts the features of ROI in the image. The abnormality or tumor can easily be detected in the segmented image.

\begin{figure}[h]
\begin{subfigure}[ht]{0.16\textwidth}
    \includegraphics[width=\textwidth]{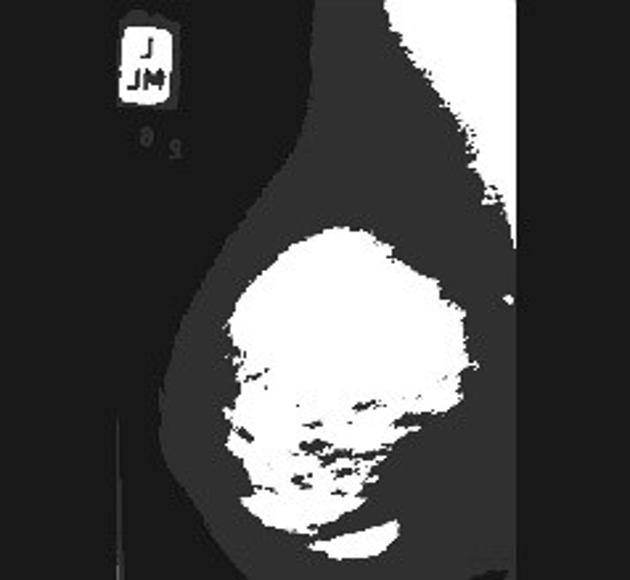}
    \caption{}
  \end{subfigure}\hfill
  \begin{subfigure}[ht]{0.16\textwidth}
    \includegraphics[width=\textwidth]{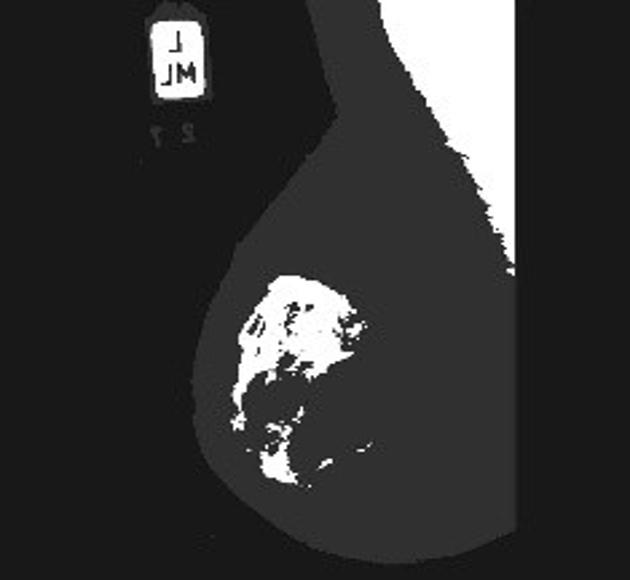}
    \caption{}
  \end{subfigure}\hfill
  \begin{subfigure}[ht]{0.16\textwidth}
    \includegraphics[width=\textwidth]{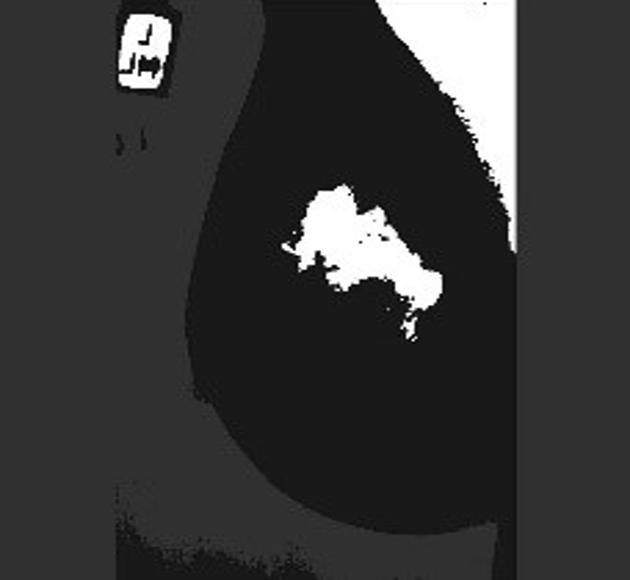}
    \caption{}
  \end{subfigure}\hfill
  \begin{subfigure}[ht]{0.16\textwidth}
    \includegraphics[width=\textwidth]{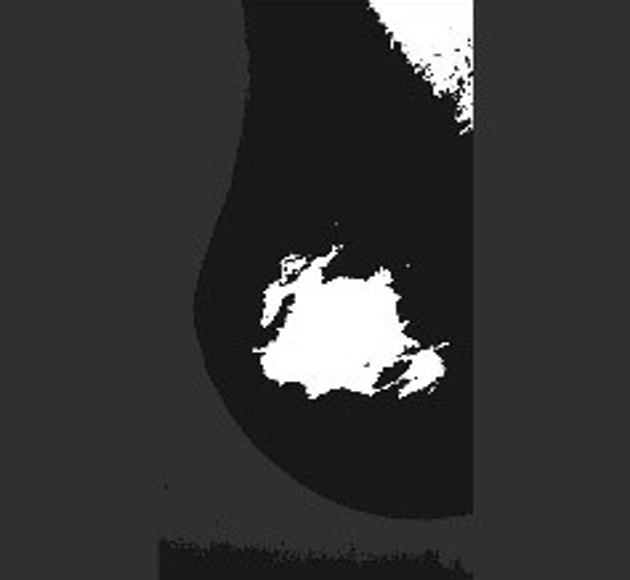}
    \caption{}
  \end{subfigure}\hfill
  \begin{subfigure}[ht]{0.16\textwidth}
    \includegraphics[width=\textwidth]{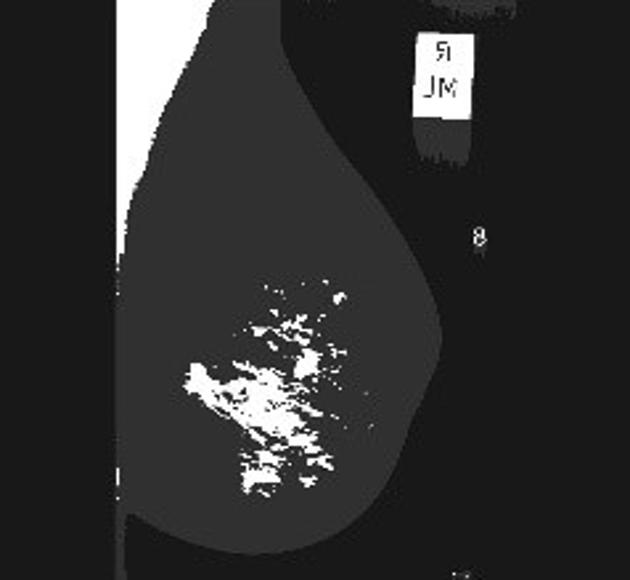}
    \caption{}
  \end{subfigure}\hfill
  \begin{subfigure}[ht]{0.16\textwidth}
    \includegraphics[width=\textwidth]{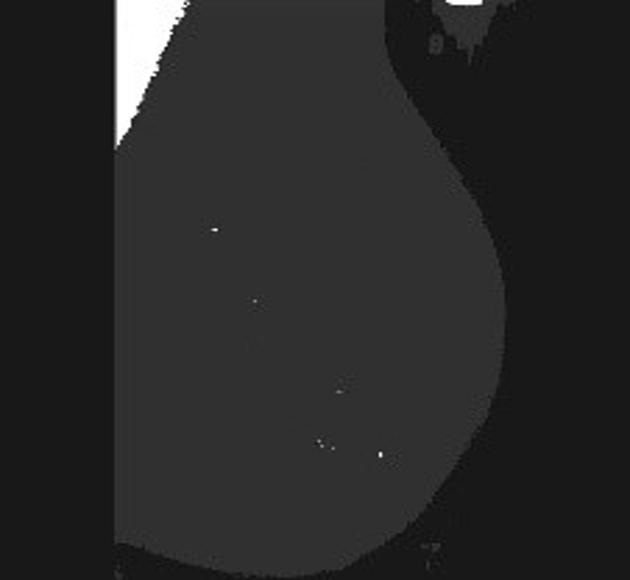}
    \caption{}
  \end{subfigure}

\caption{Sample images after segmentation (a) ARCH, (b) ASYM, (c) CALC, (d) CIRC, (e) MISC, (f) NORM.}
\label{samplesegmented}
\end{figure}

Figure \ref{samplesegmented} shows a few sample images obtained after segmentation. 

\subsection{Wavelet Transform}
Wavelet transforms of the image scans are done in the next step of our purposed network. It works as the feature extraction for our CNN working at the end to classify the breast cancer classes. Three image outputs are obtained from wavelet transform, vertical, horizontal and diagonal. Each containing the information of its respective component.

\begin{figure}[h]
\centerline{\includegraphics[height=300pt, width=.9\columnwidth]{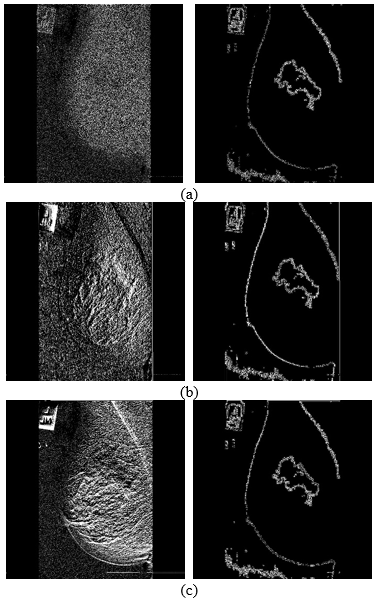}}
\caption{Sample images Level 1 Wavelet transform. (a) Diagonal, (b) Vertical, (c) Horizontal.}
\label{sampleimage}
\end{figure}

Three level wavelets transform as shown in Figure \ref{sampleimage} is performed on the original image as well as on the segmented image taken from the section above. These images are then re-sized in the size of original image to be fed into the CNN. 

\subsection{Deep Neural Network (DNN)}
\begin{figure}[H]
\centerline{\includegraphics[width=1\columnwidth]{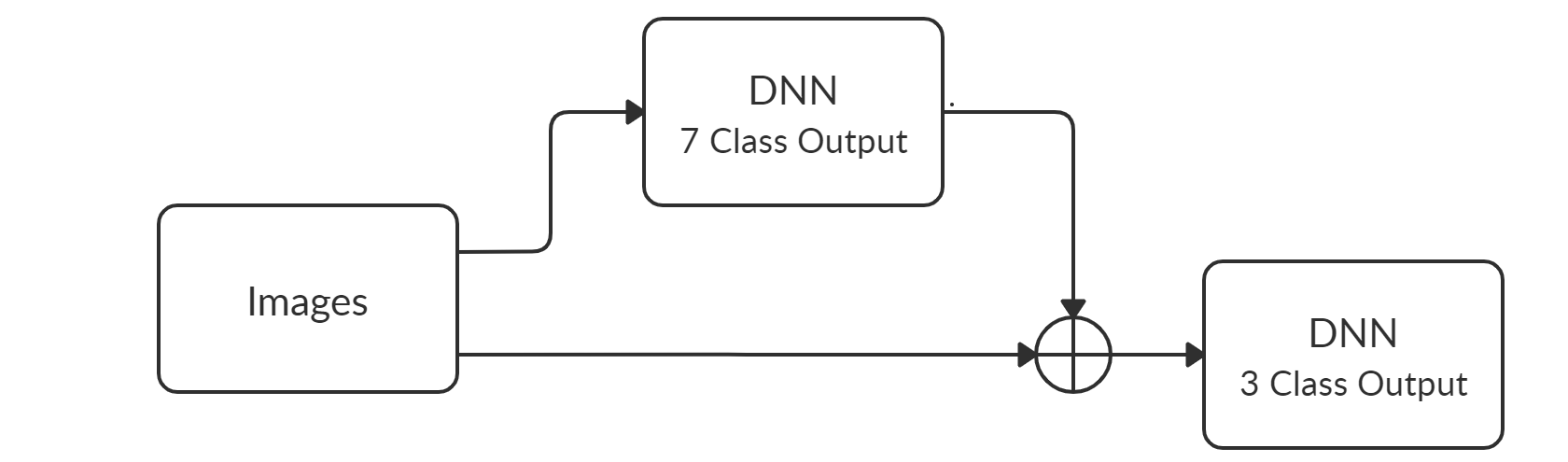}}
\caption{Proposed hybrid system of two DNNs}
\label{workflow2}
\end{figure}

DNN is used at the end of our proposed sytem for the classification among selected classes. A hybrid system of two DNNs working together is proposed. The first DNN is trained on 7 classes of abnormalities and gives its output to the second DNN which then trains on the output of first DNN along with all the images. This is a novel approach towards this kind of problem which uses the knowledge of two DNNs for classification. Figure \ref{workflow2} shows the workflow of both the networks working together.  \newline
Pre-trained networks are used with the transfer learning technique to train our networks. The weights of first few layers of the pre-trained network are kept frozen and last three layers are replaced by new modified layers which are learnable.
AlexNet, VGG, GoogleNet, ResNet50 networks are further trained on the dataset discussed.


In the training phase, hyper-parameters are to be set in order to achieve best results from the training on our dataset. These parameters include optimizers, activation function, and learning rates. Furthermore, the number of epochs are defined before the training starts. 


The above-mentioned techniques, segmentation, and wavelet transforms, are used as data augmentation and fed to the network along with original images for the training purpose. Apart from this, the basic augmentation is also done which includes the rotation, translation, scaling, and sheer of the images with random values.

The dataset is split onto two parts: the training dataset and the validation dataset. The training dataset is used to train the system by updating the weights of the layers added while the validation dataset serves as a feedback to the network after every few iterations and helps the system to improve and avoid overfitting.

\begin{table}[htbp]
\captionsetup{justification=centering, labelsep=newline}
\caption{Network Hyper-parameters}
\centering
\begin{tabular}{|c|c|}
\hline
\textbf{Parameters}& \textbf{Values} \\
\hline 
Max Epochs & 30 \\
\hline
Mini Batch Size&10\\
\hline
Initial Learning Rate&3e-4\\
\hline
Activation Function&ReLu\\
\hline
Optimizer&SGDM\\
\hline
\end{tabular}
\label{tab1}
\end{table}


\section{Results}
For the purpose of validation the results Receiver Operating Characteristic (ROC) curve is plotted for all the networks under study, it is actually a probability curve which is used as a performance measurement in a classification problems related to medical field \cite{b41}. 
For each class a separate ROC curve is plotted which have True Positive Rate (TPR) and False Positive Rate (FPR) on its x and y axis respectively \cite{b53}. Equation \ref{tpr} and \ref{fpr} below shows the formula to calculate TPR and FPR from the classification results.

\begin{equation}
    TPR = \dfrac{True Positive}{True Positive+False Negative} 
    \label{tpr}
\end{equation}
\begin{equation}
    FPR = \dfrac{False Positive}{True Negative+False Positive} 
    \label{fpr}
\end{equation}

Area under the ROC curves (AUC) tells us how the network performed for the specific class. The greater the AUC, the better the classification .

\begin{figure}[h]
\centerline{\includegraphics[width=1\columnwidth]{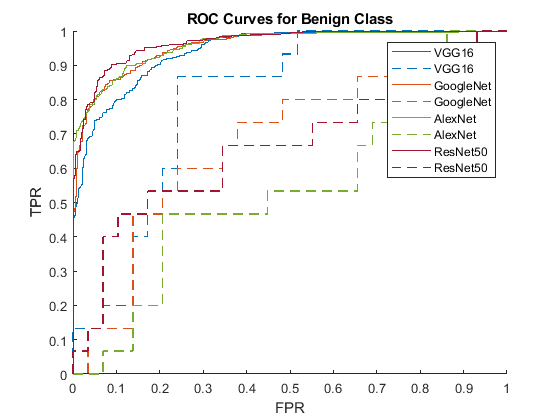}}
\caption{ROC curve for Benign class after classification}
\label{benign}
\end{figure}

\begin{figure}[h]
\centerline{\includegraphics[width=1\columnwidth]{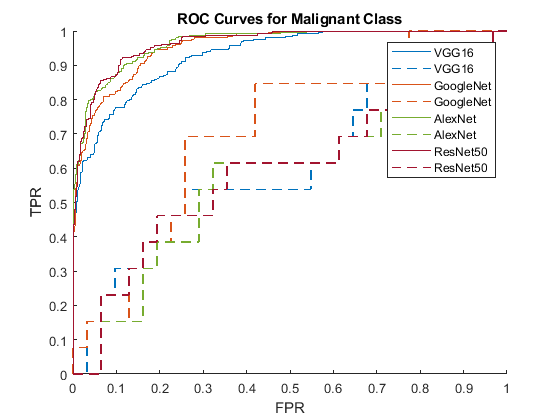}}
\caption{ROC curve for Malignant class after classification}
\label{malignant}
\end{figure}

\begin{figure}[H]
\centerline{\includegraphics[width=1\columnwidth]{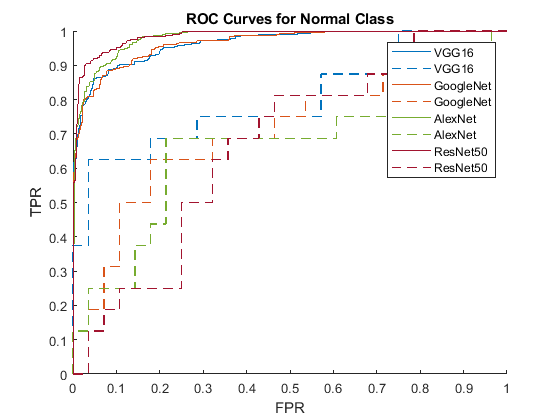}}
\caption{ROC curve for Normal class after classification}
\label{normal}
\end{figure}

In order to obtain the results, first the networks were trained using the proposed architecture on the original dataset using transfer learning with pre-trained networks. Then the same networks were trained on the images obtained after doing the segmentation and wavelet transform of the original images. The results are shown in the form of ROC curves in the Fig. \ref{benign},\ref{malignant} and \ref{normal} for the three classes which are Benign, Malignant and Normal respectively where dotted lines represent the curves of networks trained on only original images while solid lines show the curves for networks trained on segmented and wavelet transforms along with the original data. AUC of these classes are stated in the Table \ref{bbenign_t}, \ref{malignant_t} and \ref{normal_t}.




\begin{table}[h]
\centering
\captionsetup{justification=centering, labelsep=newline}
\caption{AUCs of Benign class after classification}
\begin{tabular}{|l|c|c|}
\hline
\multicolumn{1}{|c|}{\multirow{2}{*}{\textbf{\begin{tabular}[c]{@{}c@{}}Transfer Learning\\ Network\end{tabular}}}} & \multicolumn{2}{c|}{\textbf{Benign Class AUC}}                                                                             \\ \cline{2-3} 
\multicolumn{1}{|c|}{}                                                                                              & \textbf{Original Data} & \textbf{\begin{tabular}[c]{@{}c@{}}Data after Segmentation\\  and Wavelet transform\end{tabular}} \\ \hline
\textbf{VGG16}                                                                                                      & 0.800                  & 0.943                                                                                             \\ \hline
\textbf{GoogleNet}                                                                                                  & 0.685                  & 0.957                                                                                             \\ \hline
\textbf{AlexNet}                                                                                                    & 0.55                   & 0.959                                                                                             \\ \hline
\textbf{ResNet50}                                                                                                   & 0.659                  & 0.966                                                                                             \\ \hline
\end{tabular}
\label{bbenign_t}
\end{table}

\begin{table}[H]
\centering
\captionsetup{justification=centering, labelsep=newline}
\caption{AUCs of Malignant class after classification}
\begin{tabular}{|l|c|c|}
\hline
\multicolumn{1}{|c|}{\multirow{2}{*}{\textbf{\begin{tabular}[c]{@{}c@{}}Transfer Learning\\ Network\end{tabular}}}} & \multicolumn{2}{c|}{\textbf{Malignant Class AUC}}                                                                          \\ \cline{2-3} 
\multicolumn{1}{|c|}{}                                                                                              & \textbf{Original Data} & \textbf{\begin{tabular}[c]{@{}c@{}}Data after Segmentation\\  and Wavelet transform\end{tabular}} \\ \hline
\textbf{VGG16}                                                                                                      & 0.620                  & 0.930                                                                                             \\ \hline
\textbf{GoogleNet}                                                                                                  & 0.707                  & 0.954                                                                                             \\ \hline
\textbf{AlexNet}                                                                                                    & 0.573                  & 0.964                                                                                             \\ \hline
\textbf{ResNet50}                                                                                                   & 0.585                  & 0.965                                                                                             \\ \hline
\end{tabular}
\label{malignant_t}
\end{table}


\begin{table}[H]
\centering
\captionsetup{justification=centering, labelsep=newline}
\caption{AUCs of Normal class after classification}
\begin{tabular}{|l|c|c|}
\hline
\multicolumn{1}{|c|}{\multirow{2}{*}{\textbf{\begin{tabular}[c]{@{}c@{}}Transfer Learning\\ Network\end{tabular}}}} & \multicolumn{2}{c|}{\textbf{Malignant Class AUC}}                                                                          \\ \cline{2-3} 
\multicolumn{1}{|c|}{}                                                                                              & \textbf{Original Data} & \textbf{\begin{tabular}[c]{@{}c@{}}Data after Segmentation\\  and Wavelet transform\end{tabular}} \\ \hline
\textbf{VGG16}                                                                                                      & 0.796                  & 0964                                                                                              \\ \hline
\textbf{GoogleNet}                                                                                                  & 0.712                  & 0.965                                                                                             \\ \hline
\textbf{AlexNet}                                                                                                    & 0.665                  & 0.979                                                                                             \\ \hline
\textbf{ResNet50}                                                                                                   & 0.665                  & 0.985                                                                                             \\ \hline
\end{tabular}
\label{normal_t}
\end{table}


We can see from the above results consisting of ROC and AUC obtained from different transfer learning models that models trained on dataset consisting of wavelet transforms and segmented images performed significantly better on all the networks. Furthermore we also deduced that ResNet50 performs much better than than the other models under study while using transfer learning.


\subsection{Comparison with related work}
In Table \ref{comparison} we have provided overview of few related works and their performances on the Mini-MIAS dataset. However due to the variance in metrics used and the number of classes, a detailed comparison with the methods of other works is unfeasible. Most of the previous studies have done two class classification to mark the presence or absence of cancer tissues and have acquired some promising results. \cite{b50}.


\begin{table}[H]
\centering
\centering
\captionsetup{justification=centering, labelsep=newline}
\caption{Comparison of classification with related work based on AUC.}
\begin{tabular}{|l|l|l|l|}
\hline
\multicolumn{1}{|c|}{\textbf{Paper}}                           & \multicolumn{1}{c|}{\textbf{Method}}                                                          & \multicolumn{1}{c|}{\textbf{Dataset}}                     & \multicolumn{1}{c|}{\textbf{AUC}}                                                              \\ \hline
\begin{tabular}[c]{@{}l@{}}Rouhi et al. \\ (2015)\cite{b52}\end{tabular} & SNN                                                                                           & \begin{tabular}[c]{@{}l@{}}Mini-MIAS,\\ DDSM\end{tabular} & 0.92 (Binary)                                                                                  \\ \hline
\begin{tabular}[c]{@{}l@{}}Valarmathie\\ (2016) \cite{b42}\end{tabular}   & Fuzzy + ANN                                                                                   & Mini-MIAS                                                 & 0.99 (Binary)                                                                                  \\ \hline
\begin{tabular}[c]{@{}l@{}}Rabidas\\ (2016)\cite{b43}\end{tabular}       & DRLBP, DRLTP                                                                                  & Mini-MIAS                                                 & 0.98 (Binary)                                                                                  \\ \hline
\begin{tabular}[c]{@{}l@{}}Jaffar\\ (2017)\cite{b51}\end{tabular}        & CNN (COM)                                                                                     & \begin{tabular}[c]{@{}l@{}}Mini-MIAS,\\ DDSM\end{tabular} & 0.92 (Binary)                                                                                  \\ \hline
\begin{tabular}[c]{@{}l@{}}Proposed\\ Method\end{tabular}      & \begin{tabular}[c]{@{}l@{}}Wavelet Transform\\ Segmentation \\ Transfer Learning\end{tabular} & Mini-MIAS                                                 & \begin{tabular}[c]{@{}l@{}}(3 Class)\\ 0.97 Benign\\ 0.97 Malignant\\ 0.99 Normal\end{tabular} \\ \hline
\end{tabular}
\label{comparison}
\end{table}


In a comparison of the best results through our method with those found in related work, it is clear that our method is outstanding for all three classes with balanced values.

\section{Conclusions}
To summarize, this study effectively implemented the CNN on mammograms for the detection of abnormal masses in breasts which cause cancer. Different CNN architectures are used to get the state of the art accuracies on the mini-MIAS dataset. Hence, the study shows the significance of data augmentation before feeding the images to the network for training.

Segmentation and wavelet transforms help the network in extracting the important features from the mammogram. The study's main contribution, that is the combination of transfer learning along with wavelet transform, significantly improves the results by using segmentation and wavelet transforms as pre-processing for the images before the training process starts. The combination of segmentation and wavelet transform improves the overall efficiency of CNN by helping in extracting features meanwhile the unique hybrid system that uses transfer learning alongside wavelet transforms and segmentation greatly affects the resulting ROC curves positively. Segmentation gives the information about edges of tumorous regions while wavelet transform helps in localization in frequency domain which then gives the information of tumorous regions in the form of contours in images.

The betterment in results is clear considering the AUC-ROC curve achieves is greater than 0.95 on all three classes which is considerably good considering three-class classifications have not been measured using AUC-ROC before.

The dataset utilized for this paper was fairly small and yielded favorable results, therefore the system can be tested on bigger datasets comprising of a more huge number of elements to further test its accuracy. One estimated hurdle could be the application of transfer learning to bigger datasets however that poses a suitable foundation for future work.

\ifCLASSOPTIONcaptionsoff
  \newpage
\fi



%

%
\begin{IEEEbiography}[{\includegraphics[width=1in,height=1.25in,clip,keepaspectratio]{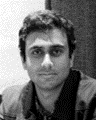}}]{Ahmed Rasheed} earned his Bachelor's degree in Electrical Engineering, majoring in Electrical Power, from Air University Islamabad, Pakistan. He is currently an Electrical Engineering Master's student majoring in Digital System and Signal Processing, at School of Electrical Engineering and Computer Science (SEECS) at the National University of Sciences and Technology (NUST) Islamabad, Pakistan.  He is also serving as a Research Assistant at the laboratory of Adaptive Signal Processing (ASP) at SEECS, specializing in handling machine learning, deep learning, time series prediction and biomedical image processing tasks.
\end{IEEEbiography}

\begin{IEEEbiography}[{\includegraphics[width=1in,height=1.25in,clip,keepaspectratio]{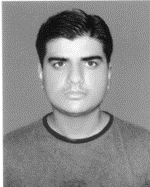}}]{Muhammad Shahzad Younis} received the bachelor’s degree from National University of Sciences and Technology, Islamabad, Pakistan, in 2002, the master’s degree from the University of Engineering and Technology, Taxila, Pakistan, in 2005, and the PhD degree from University Technology PETRONAS, Perak, Malaysia in 2009, respectively. Before joining National University of Sciences and Technology (NUST), he was Assistant Manager at a research and development organization named AERO where he worked on different signal processing and embedded system design applications.  He is currently working as an Assistant professor in the Department of Electrical Engineering in School of Electrical Engineering and Computer Science (SEECS)-NUST. He has published more than 25 papers in domestic and international journals and conferences. His research interests include Statistical Signal Processing, Adaptive Filters, Convex Optimization Biomedical signal processing, wireless communication modelling and digital signal processing. 
\end{IEEEbiography}

\begin{IEEEbiography}[{\includegraphics[width=1in,height=1.25in,clip,keepaspectratio]{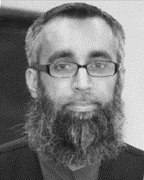}}]{Junaid Qadir} (SM’ 14) completed his BS in Electrical Engineering from UET, Lahore, Pakistan and his PhD from University of New South Wales, Australia in 2008. He is currently an Associate Professor at the Information Technology University (ITU)-Punjab, Lahore, Pakistan. He is the Director of the IHSAN Lab at ITU that focuses on deploying ICT for development, and is engaged in systems and networking research. Prior to joining ITU, he was an Assistant Professor at the School of Electrical Engineering and Computer Sciences (SEECS), National University of Sciences and Technology (NUST), Pakistan. At SEECS, he directed the Cognet Lab at SEECS that focused on cognitive networking and the application of computational intelligence techniques in networking. He has been awarded the highest national teaching award in Pakistan—the higher education commission’s (HEC) best university teacher award—for the year 2012-2013. He has been nominated for this award twice (2011, and 2012-2013). His research interests include the application of algorithmic, machine learning, and optimization techniques in networks. In particular, he is interested in the broad areas of wireless networks, cognitive networking, software-defined networks, and cloud computing. He is a regular reviewer for a number of journals and has served in the program committee of a number of international conferences. He serves as an Associate Editor for IEEE Access, IEEE Communication Magazine, and Springer Nature Big Data Analytics. He was the lead guest editor for the special issue “Artificial Intelligence Enabled Networking” in IEEE Access and the feature topic “Wireless Technologies for Development” in IEEE Communications Magazine. He is a member of ACM, and a senior member of IEEE.
\end{IEEEbiography}

\begin{IEEEbiography}[{\includegraphics[width=1in,height=1.25in,clip,keepaspectratio]{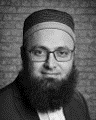}}]{Muhammad Bilal} is Associate Professor of Big Data and Artificial Intelligence (AI) at Big Data Laboratory, University of the West of England (UWE), Bristol. He holds a PhD in Big Data Analytics from UWE, Bristol. He has multi-disciplinary research interests that span across Intelligent Systems, Internet of Things (IoTs), AI Product Design, Digital Health, Visual Analytics, GIS and Semantic technologies. Dr Bilal has led the development of various enterprise-grade software projects ranging from financials to healthcare. He has vast expertise in collaborative research design and execution. So far, he has completed research and development projects of £3.7 Million at Big Data Lab in collaboration with leading UK businesses. He has also authored more than 50 research articles at high-impact journals and international conferences.
\end{IEEEbiography}






\end{document}